\pdfoutput=1

\documentclass[11pt]{article}

\usepackage[]{ACL2023}

\usepackage{times}
\usepackage{latexsym}
\usepackage{soul}
\usepackage{url}
\usepackage{amsmath}
\usepackage{amsthm}
\usepackage{booktabs}
\usepackage{algorithm}
\usepackage{algorithmic}
\usepackage[switch]{lineno}
\usepackage{graphicx}
\usepackage[inkscapelatex=false]{svg}
\usepackage{subfigure}
\usepackage{makecell}
\usepackage{multirow}
\usepackage{xcolor}
\usepackage{amssymb}
\usepackage{hyperref}

\usepackage[T1]{fontenc}
\usepackage[utf8]{inputenc}

\usepackage{microtype}
\usepackage{inconsolata}

\title{Enhancing Dialogue Generation via Dynamic Graph Knowledge Aggregation}

\author{Chen Tang\textsuperscript{1}, Hongbo Zhang\textsuperscript{2}, Tyler Loakman\textsuperscript{2},  Chenghua Lin\textsuperscript{2}\footnotemark[1]  ~and Frank Guerin\textsuperscript{1}\\
    Department of Computer Science, The University of Sheffield, UK \\
    Department of Computer Science, The University of Surrey, UK \\
  \texttt{\{chen.tang,f.guerin\}@surrey.ac.uk} \\
    \texttt{\{hzhang183,tcloakman1,c.lin\}@sheffield.ac.uk} }
  
\begin{document}
\maketitle

\renewcommand{\thefootnote}{\fnsymbol{footnote}} 
\footnotetext[1]{Corresponding author.} 
\renewcommand*{\thefootnote}{\arabic{footnote}}

\begin{abstract}
Incorporating external graph knowledge into neural chatbot models has been proven effective for enhancing dialogue generation. However, in conventional graph neural networks (GNNs), message passing on a graph is independent from text, resulting in the graph representation hidden space differing from that of the text. This training regime of existing models therefore leads to a semantic gap between graph knowledge and text. In this study, we propose a novel framework for knowledge graph enhanced dialogue generation. We dynamically construct a multi-hop knowledge graph with pseudo nodes to involve the language model in feature aggregation within the graph at all steps. To avoid the semantic biases caused by learning on vanilla subgraphs, the proposed framework applies hierarchical graph attention to aggregate graph features on pseudo nodes and then attains a global feature. Therefore, the framework can better utilise the heterogeneous features from both the post and external graph knowledge. Extensive experiments demonstrate that our framework outperforms state-of-the-art (SOTA) baselines on dialogue generation. Further analysis also shows that our representation learning framework can fill the semantic gap by coagulating representations of both text and graph knowledge. Moreover, the language model also learns how to better select knowledge triples for a more informative response via exploiting subgraph patterns within our feature aggregation process. Our code and resources are available at \url{https://github.com/tangg555/SaBART}.
\end{abstract}

\section{Introduction}
Recent years have seen a surge of interest in developing chatbots with the facilitation of large-scale knowledge \cite{tang2022recent}. As a highly expressive data format, Knowledge Graphs (e.g. ConceptNet and DBpedia), which include world facts, are considered to be a key factor in building an effective dialogue generation system \cite{zhou2018commonsense}. In order to incorporate graph-structured knowledge, a range of Graph Neural Networks such as Graph Attention Networks (GATs)~\cite{velickovic2017graph,brody2021attentive} and Graph Convolutional Networks (GCNs)~\cite{kipf2016semi} have been proposed to learn representations of the topological structure of the knowledge graph via message passing between entities. In open-domain dialogue generation, these GNNs are further embedded into generative frameworks to feed graph knowledge features into the language models (LMs). 

\begin{figure}[t]
\centering
\includegraphics[width=0.9\columnwidth]{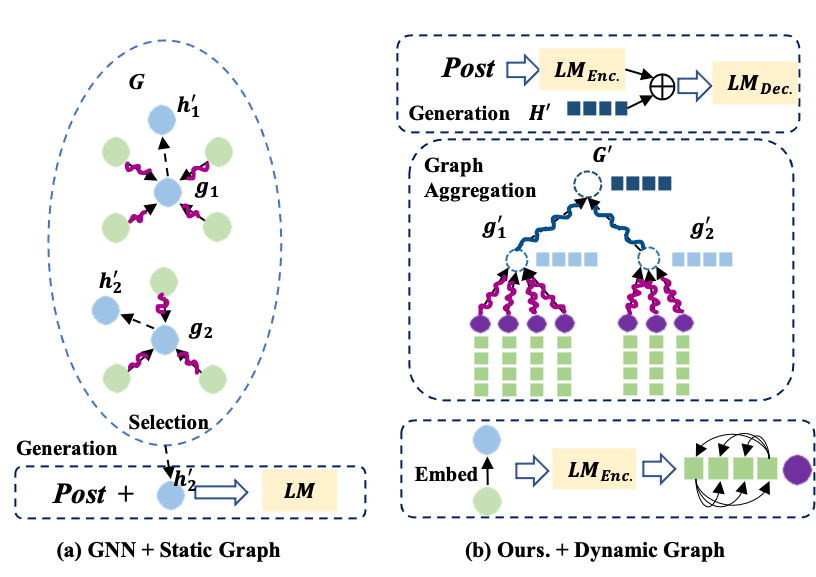}
\caption{Conventional GNNs vs. Ours.} 
\label{fig:intro}
\end{figure}

Despite prior success in leveraging graph knowledge with graph neural networks (GNN) \cite{zhou2018commonsense,zhang-etal-2020-grounded}, current generative frameworks are still hindered by the representation gap in the hidden space between the LMs and GNNs, which poses significant challenges in exploiting graph knowledge in the subsequent text decoding process. As illustrated in \autoref{fig:intro}, prior works using GNNs \cite{Copynet,ghazvininejad2018knowledge,zhou2018commonsense,zhang-etal-2020-grounded} tend to fuse the graph features by transforming them into text form and then feeding them into the language model, which acts as a ``copy'' mechanism. In other words, these networks run as a pipeline where the graph knowledge is firstly transformed into additional text to avoid the problem of language model encoding brought about by the heterogeneous graph features. However, these separate encoding stages result in neural networks learning suboptimal representations of graph knowledge, which leads to information loss. With large-scale pretrained models such as GPT-2~\cite{radford2019language}, BART~\cite{lewis-etal-2020-bart} and T5~\cite{raffel2020exploring} being widely adopted in recent advances in dialogue generation, the drawbacks that arise from incompatibility between GNNs and LMs becomes a more severe problem, prohibiting chatbot systems from leveraging graph structured data effectively.

In order to address the aforementioned challenges, we propose a novel representation learning framework to facilitate language understanding and generation,  which permits effective incorporation of heterogeneous features via a dynamic graph knowledge aggregation mechanism.  In contrast to existing works \cite{Copynet,ghazvininejad2018knowledge,zhou2018commonsense,zhang-etal-2020-grounded} which incorporate graph knowledge with conventional GNNs (causing inadequacies in representation learning), we propose to involve language models in both text and graph knowledge incorporation at all steps via hierarchically aggregating knowledge on a dynamic pseudo graph. During the knowledge aggregation process, knowledge triples are reorganised as shown in \autoref{fig:intro} (b), where pseudo nodes are created to learn conceptual representations from original knowledge triples. Conceptual semantics are forced to coagulate into pseudo nodes, and finally merge into a condensed feature vector to fill the semantic gap of the encoded text features. Our approach for incorporating text and graph knowledge features can be adapted to all language models with an encoder-decoder architecture. In this study, we choose BART~\cite{lewis-etal-2020-bart}, a SOTA language model for generation, as our language model in our experiments. This framework will hereinafter be referred to as SaBART (Subgraph-Aggregation BART).

During subgraph knowledge aggregation, the language model is involved in learning three levels of features: (1) Subword-level, where conceptual embeddings are connected to entity mentions within text; (2) Knowledge-level, where original triples are transformed by language encoding; and (3) Semantic-level, where the context vector encoded from text is involved in knowledge aggregation. This implies that the neural networks are able to access both the text and graph features during representation learning. The text and graph unified encoding process also avoids the information loss caused by the representation shift in vanilla GNNs, thus improving efficiency and efficacy. Extensive experiments demonstrate that our proposed framework significantly outperforms current SOTA baselines in dialogue generation. We also conduct in-depth analysis into the underlying mechanism of why our proposed approach better incorporates heterogeneous features. Our contributions can be summarised as follows:
\begin{itemize}
\item We propose a novel representation learning framework where graph and text features can be effectively aggregated via hierarchical knowledge aggregation on a dynamically constructed pseudo graph;
\item We conduct a comprehensive set of experiments to demonstrate the effectiveness of our proposed approach, where our framework achieves SOTA performance on the commonsense knowledge graph enhanced dialogue generation dataset;
\item We conduct in-depth experiments to analyse the improvement of representation learning on both graph and text knowledge, and investigate the mechanism to address this representation gap problem of learning heterogeneous features. 
\end{itemize}

\begin{figure*}[t]
\centering
\includegraphics[width=0.92\linewidth]{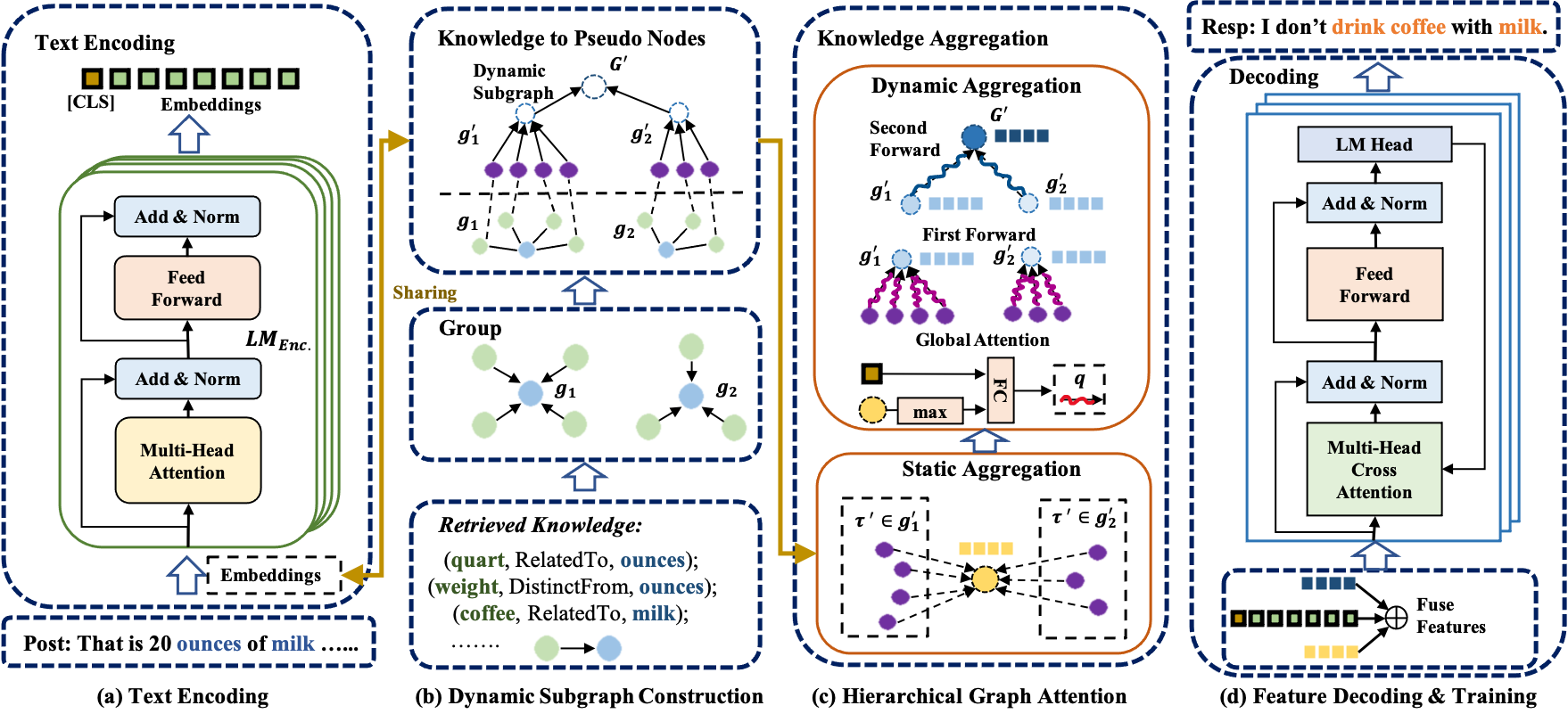}
\caption{The overall framework of SaBART. We split the whole framework into four parts, which contain co-dependencies. For instance, the concepts and relations will create new sub-word embedding placeholders in language models after flattening into text sequences to make the graph and language models share low-dimensional representations.} 
\label{fig:overview}
\end{figure*}

\section{Related Works}
In this section, we introduce related works by summarising recent advances in the knowledge enhanced dialogue generation task, as well as the SOTA approaches for injecting graph knowledge into generative frameworks.

\paragraph{Knowledge Enhanced Dialogue Generation}
As a data-driven approach, deep learning based chatbots rely on access to large amounts of knowledge~\cite{zhao-etal-2020-knowledge-grounded} to generate interesting and informative dialogues like humans. In order to realise a commonsense-aware and semantic-aware chatbot, more and more studies~\cite{yu-etal-2022-diversifying-content,huang-etal-2022-improving,tang2022terminology,zhao2023evaluating} aim to consider external knowledge beyond pure dialogue exchanges to facilitate generation, where knowledge graphs containing topological structural knowledge are an important research direction to facilitate logical reasoning. In this study, we aim to improve the performance of neural frameworks through using additional knowledge graphs as external resources. Therefore, studies with external knowledge other than knowledge graphs, such as content planning~\cite{tang-etal-2022-ngep}, retrieved documents~\cite{yu-etal-2022-diversifying-content}, and mixed resources~\cite{wu2022improving}, are not directly compared in this paper. We believe our learning pattern for handling heterogeneous features~\cite{zhanga2023cadge} can provide inspiration to other types of knowledge grounded conversational systems as well.

\paragraph{Injecting Graph Knowledge into Generative Frameworks}
With pre-training techniques being widely adopted, large-scale language models such as UniLM \cite{dong2019unified}, GPT-2 \cite{radford2019language}, and BART~\cite{lewis-etal-2020-bart,loakman2023twistlist} have become the base models for various dialogue generation systems. However, these language models generally input only sequence formatted text, and cannot directly incorporate features of graph knowledge. Usually graph knowledge has to firstly be flattened into a sequence of tokens~\cite{tang2022etrica}, or encoded as feature vectors~\cite{zhao-etal-2020-knowledge-grounded} before being fed into a language model. \citet{zhou2018commonsense} uses GRUs and two graph attention modules to select appropriate triples to incorporate into responses. In order to exploit the benefits of multi-hop knowledge, \citet{zhang-etal-2020-grounded} adds an attention mechanism in a similar way to filter the appropriate knowledge. Finally, \citet{tuan2019dykgchat} propose a model which selects the output from a sequence-to-sequence model and a multi-hop reasoning model at each time step.


\section{Methods}
As illustrated in \autoref{fig:overview}, our approach aims to improve dialogue generation of language models by better incorporating heterogeneous features with an effective knowledge aggregation framework on retrieved knowledge triples.

\subsection{Task Definition}
In our task, the given input includes a textual post $ X = \{x_1, x_2, ..., x_n\}$ 
where $x_n$ denotes the $n$-th token, and a graph knowledge base $ G = \{\tau_1, \tau_2, ..., \tau_k\}$. $\tau$ denotes a triple $ \{h, r, t\}$, where $h$, $r$, and $t$ refer to the head entity, the relation, and the tail entity, respectively. These triples represent the entities contained in the posts and reference responses, and the relations between them. Provided with these two kinds of input, the generation model is required to generate a response $ Y = \{y_1, y_2, ..., y_m\} $ by modeling the conditional probability distribution $ P(Y|X, G)$.

\subsection{Dynamic Subgraph Construction} \label{sec:dynamic-subgraph} 
The graph knowledge is obtained by retrieving concept triples from ConceptNet, which are contained in the posts and reference responses.
Our knowledge retrieval process is implemented by word matching (concepts in ConceptNet take the form of single words) and rule filtering to collect knowledge triples, which resembles the strategy of \citet{zhou2018commonsense}. 
This process involves recognising relevant conceptual entities contained in the input post, and retrieving directly connected concept entities in the responses, with the goal of exploiting these entities in output responses.\footnote{The concepts and their relations come from ConceptNet \url{https://conceptnet.io/}.} Therefore, during knowledge retrieval, the retrieved knowledge triples are grouped according to the mentions of conceptual entities from the posts. For example, the post given in \autoref{fig:overview} has ``milk'' recognised as an entity mention, which in turn retrieves relevant triples, e.g. (\texttt{coffee}, \texttt{RelatedTo}, \texttt{milk}). First of all, we group the retrieved knowledge triples as follows:
\begin{align}
    \text{ent}_{1}, \text{ent}_{2}, ...,\text{ent}_{n} \in X \cup G \\
    g_i = \{\tau_1, \tau_2, ...,\tau_j\}\quad \mathrm{s.t.}~\text{ent}_j \in \tau
\end{align}
where triples $\tau$ containing the same conceptual entity $ent_i$ are grouped as a subgraph $g_i$ for the post. In contrast to existing works~\cite{Copynet,ghazvininejad2018knowledge,zhou2018commonsense,zhang-etal-2020-grounded} that encode knowledge and select triples on separate subgraphs (leading to biased and incomplete feature learning), we propose to reconstruct $G$ with pseudo nodes, so that pseudo nodes can dynamically connect each $g_i$ to form a global graph:
\begin{align}
    pseu_{\tau_j} & =  \text{LM}_{emb}(F_{flatten}(\tau_j)) \\
    F_{flatten}(\tau_j) & = [x^h_j, x^{r_0}_j, x^{r_1}_j, x^t_j] \\
    F_{insert}(F_{flatten} & (\tau_j)|\tau_j \in G)  \stackrel{emb}{\longrightarrow } \text{LM}_{emb} 
\end{align}
where $F_{flatten}$ flattens the triple of $(h, r, t)$ to a text sequence, e.g., (coffee, RelatedTo, milk) will be flattened to ``coffee related to milk''. In ConceptNet, $h$ and $t$ consist of single words, and $r$ is the relation of the two words. These words are transformed into BPE (byte-pair encoding) pieces~\cite{lewis-etal-2020-bart}, and distinctively inserted into the sub-word embedding layer of the LM. This is performed in order to learn the semantics of both the entity mentions in the post, as well as the topological structure of the graph $pseu_{\tau_j} \in \mathbb{R}^{WE}$ ($E$ denotes the size of the word embeddings), which constitutes the representation of the whole triple. We replace the original triples of $G$ with the basic pseudo nodes (in purple)
, where $W$ is the word length of flattened $\tau_j$, and $C$ is the embedding size of the LM. 
On top of the basic pseudo nodes, hierarchical pseudo layers are created to connect all subgraphs:
\begin{align}
    g^{'}_i & = \{\tau^{'}_1, \tau^{'}_2, ...,\tau^{'}_j\}\quad \mathrm{s.t.}~\tau_j \in g_i \label{eq-g_i} \\ 
    \tau^{'}_j & = \{pseu_{\tau_j}, r_a, pseu_{g_i}\}
\end{align}
where $g^{'}_i$ denotes the subgraph rebuilt by pseudo nodes $pseu_{\tau_{1<j<=|g_i|}} \in \mathbb{R}^{WC}$. They are connected to $pseu_{g_i}$ with a relation $r_a$, whose weight is calculated with an attention mechanism introduced in Sec.~\ref{KA}.
\begin{align}
    G^{'} & = \{T_1, T_2, ...,T_k\} \label{eq-G}\\ 
    T_k & = \{pseu_{g_k}, r_a, pseu_{G}\}
\end{align}
Here $pseu_{G} \in \mathbb{R}^{WC}$ is the root node representing the features of the whole graph $G^{'}$. $pseu_{G}$ as the new pseudo knowledge graph is the set of the aforementioned pseudo triples transformed from the original triples. 

\subsection{Hierarchical Knowledge Aggregation}\label{KA}
Instead of learning graph features by message passing on graph nodes, we implement a novel representation learning framework, where we train the neural network to learn the global features from the whole graph by hierarchically aggregating features through pseudo nodes as shown in \autoref{fig:overview}(c). Firstly, we encode features of the post to obtain a semantic vector $\mathbf{H}^{CLS} \in \mathbb{R}^{1 \times E}$, and the embeddings of input tokens  $\mathbf{H}^{X}$:
\begin{align}
   LM_{enc}(X) &= [\mathbf{H}^{CLS}; \mathbf{H}^{X}] \label{eq:E-post} \nonumber \\
    & = [emb^c_0; emb_1, emb_2, ...] 
\end{align}
where the context information of the post $\mathbf{H}^{CLS}$ will be used as context from the post, and involved in aggregating features on graph knowledge (as the query vector in the attention mechanism). Subsequently, a series of feature incorporation procedures will be conducted on our constructed graph of pseudo nodes. 

\subsubsection{Aggregation on Static Graphs}
In \S\ref{sec:dynamic-subgraph}, the original retrieved triples have been transformed into the set of pseudo nodes $pseu_{\tau_j}$. To obtain the representation of the graph, we directly aggregate the node features by calculating their mean value, which is inspired by the work of \citet{tang2021normcg} in calculating global representations.
\begin {align}
    \epsilon &= \frac{\sum_{\tau_j \in G} pseu_{\tau_j}}{|G|}
\end{align}
where $\epsilon \in \mathbb{R}^{WE}$ denotes the semantic representation of all triples. Because every node has the same weight when contributing to the global representation, $\epsilon$ will be carried into the following dynamic graph aggregation to obtain a better graph representation feature for response generation.

\subsubsection{Aggregation on Dynamic Graphs}
The aggregation process on the dynamic knowledge graph has a sequential forward procedure as follows.
\paragraph{First Forward Layer.}
In the first step, we calculate the features of $pseu_{g_i} \in g_i$:
\begin {align}
    \text{Update}(pseu_{g_i}) &= \sum_{j=1}^{|g^{'}_i|} a_{ji}^{g^{'}_i} pseu_{\tau_j} \\
    a_{ji}^{g^{'}_i} & = \frac{\exp(\beta_{ji}^{g^{'}_i})}{\sum_{j=1}^{|g^{'}_i|} \exp(\beta_{ji}^{g^{'}_i})} \label{eq-fa} \\
    \beta_j^{g^{'}_i} & = \mathbf{W}^{g^{'}_i}{[pseu_{\tau_j}; q]}^{\mathrm{T}} \\
    q &= FC([\mathbf{H}^{CLS}; \epsilon^{'}]) \\
    \epsilon^{'} &= max_{pool}(\epsilon) 
\end{align}
where $\tau^{'}_j \in g^{'}_i$ all include $pseu_{g_i}$ as the tail node (cf. \autoref{eq-g_i}); $i$ denotes the $i$-th entity mention in the post; and $j$ denotes the $j$-th triple related to the mention. $\mathbf{W}^{g^{'}_i} \in \mathbb{R}^{1 \times (W+1)E}$ is a trainable parameter multiplying the concatenation of the node representation and the context feature vector. $a_{ji}^{g^{'}_i}$ is an attention score to aggregate features of $pseu_{\tau_j}$ into the updated $pseu_{g_i}$. $q \in \mathbb{R}^{2E}$ is the query vector of the attention mechanism. $FC$ is a fully connected neural network, and $max_{pool}$ is the max-pool function transforming $\epsilon \in \mathbb{R}^{WE}$ to $\epsilon^{'} \in \mathbb{R}^{E}$.

\paragraph{Second Forward Layer.}
Similarly, when our model attends to $G^{'}$, we update the features with $pseu_{g_k}$ obtained in the first step:
\begin {align}
    \text{Update}(pseu_{G}) &= \sum_{k=1}^{|G^{'}|} a_{k}^{G^{'}} pseu_{g_k} \\
    a_{k}^{G^{'}} & = \frac{\exp(\beta_{k}^{G^{'}})}{\sum_{j=1}^{|G^{'}|} \exp(\beta_{k}^{G^{'}})} \\
    \beta_k^{G^{'}} & = \mathbf{W}^{G^{'}}{[pseu_{g_k}; q]}^{\mathrm{T}}
\end{align}
where $\mathbf{W}^{G^{'}} \in \mathbb{R}^{1 \times (W+1)E}$ is a trainable parameter, and the final $pseu_{G}$ represents the global features aggregated by $G$. The feature vector $q$ is the same as the one in the first forward layer, which acts as the global context of both the post and static knowledge graph.

\subsection{Inference and Training}
To auto-regressively generate responses, the language model predicts each token $y_t$ at time step $t$:
\begin{align}
    Y_t &= [y_1, y_2, ..., y_t]\quad \mathrm{s.t.}\quad t>0 \\
    p_{y_t}& = \text{softmax}(\mathbf{H}^{dec}\mathbf{W}^{res}) \\
    \mathbf{H}^{dec} &= \text{LM}_{dec}(\mathbf{H}^{enc}, Y_{t-1}) \\
    \mathbf{H}^{enc} &= [pseu_G;\epsilon;\mathbf{H}^{CLS};\mathbf{H}^{X}] \label{eq-H-enc}
\end{align}
where $\mathbf{H}^{enc} \in \mathbb{R}^{L \times E}$ and $\mathbf{H}^{dec} \in \mathbb{R}^{1 \times E}$ are outputs of encoders and decoders; and $L$ denotes the size of the concatenated feature vector. The dimension of $pseu_G$ here is transformed to $\mathbb{R}^{W \times E}$, so that $pseu_G$, $\mathbf{H}^{CLS}$ and $\mathbf{H}^{X}$ can be concatenated at the first dimension. $W^{res} \in \mathbb{R}^{L \times E}$ is a trainable parameter denoting the LM head in \autoref{fig:overview}(d); and $E$ denotes the size of the word embeddings. Finally, we train the whole framework with the loss function as follows:
\begin{equation}
     \mathcal{L} = - \frac{1}{N}\sum_{n=1}^N \log P(Y|X, G)
\end{equation}
where $N$ denotes the size of the test data, and $\mathcal{L}$ is the cross entropy of predicted response tokens and those of the golden responses. 

\begin{table}[tb]
\scriptsize
    \centering
    \begin{tabular}{l|ccc}
    \toprule
\textbf{Datasets}&\textbf{Train}&\textbf{Val}&\textbf{Test}\\
\midrule
\textbf{Conversational \# Pairs}& 3,384,185 & 20,000 & 10,000 \\
\textbf{Vocabulary Size} & 39,674 & 27,115 & 18,036 \\
\midrule
\textbf{Retrieved \# Entities} & 108,410 & 27,135 & 20,125\\
\textbf{Retrieved \# Triples} & 120,848 & 110,952 & 86,940 \\
\midrule
\textbf{Avg. \# Entities in Input Posts} & 92.25 & 94.64 & 92.84 \\
\textbf{Avg. \# Entities in Output Responses} & 2.33 & 2.31 & 2.33 \\
\textbf{Avg. \# Subgraphs in \# Pairs} & 5.77 & 6.47 & 5.81 \\
\textbf{Avg. \# Triples in \# Pairs} & 105.43 & 108.12 & 106.04 \\
\bottomrule
    \end{tabular}
    \caption{Data statistics of the commonsense dialogue generation dataset. Retrieved entities and triples show the unique entities and triples contained in the dataset. The definition of a subgraph refers to \autoref{sec:dynamic-subgraph}}
    \label{tab:data_stat}
\end{table}

\begin{table*}[t]
\centering \small
\resizebox{0.98\linewidth}{!}{
\begin{tabular}{l|cccc|cccc|c}
    \toprule
    \textbf{Model} & \textbf{BLEU-1} & \textbf{BLEU-2} & \textbf{BLEU-3} & \textbf{BLEU-4} & \textbf{NIST-1} & \textbf{NIST-2} & \textbf{NIST-3} & \textbf{NIST-4} & \textbf{METEOR} \\
    \hline
    \textbf{Seq2Seq} & 0.1702 & 0.0579 & 0.0226 & 0.0098 & 1.0230 & 1.0963 & 1.1056 & 1.1069 & 0.0611 \\
    \textbf{MemNet} & 0.1741 & 0.0604 & 0.0246 & 0.0112 & 1.0975 & 1.1847 & 1.1960 & 1.1977 & 0.0632  \\
    \textbf{CopyNet} & 0.1589 & 0.0549 & 0.0226 & 0.0106 & 0.9899 & 1.0664 & 1.0770 & 1.0788 & 0.0610  \\
    \textbf{CCM} & 0.1413 & 0.0484 & 0.0192 & 0.0084 & 0.8362 & 0.9000 & 0.9082 & 0.9095 & 0.0630  \\
    \textbf{UniLM} & 0.2019 & 0.0730 & 0.0305 & 0.0138 & 1.3562 & 1.4919 & 1.5082 & 1.5101 & 0.0796  \\
    \textbf{ConceptFlow} & 0.2451 & 0.1047 & 0.0493 & 0.0246 & 1.6137 & 1.7956 & 1.8265 & 1.8329 & 0.0942 \\
    \midrule
    \textbf{SaBART (ours)} & \textbf{0.3298} & \textbf{0.2113} & \textbf{0.1467} & \textbf{0.0945} & \textbf{2.9226} & \textbf{3.8386} & \textbf{4.0763} & \textbf{4.1121} & 0.1674 \\
    \textbf{- w/o dy-agg} & 0.2967 & 0.1909 & 0.1322 & 0.0846 & 2.3889 & 3.1635 & 3.3618 & 3.3914 & 0.1647  \\
    \textbf{- w/o st-agg} & 0.2927 & 0.1880 & 0.1299 & 0.0831 & 2.3712 & 3.1593 & 3.3623 & 3.3928 & \textbf{0.1684} \\
    \textbf{- w/o kg} & 0.1446 & 0.0578 & 0.0285 & 0.0155 & 1.0381 & 1.1653 & 1.2245 & 1.2327 & 0.0931 \\
    \bottomrule
\end{tabular}
}
\caption{\label{tab:reference-metrics}
Automatic evaluation on referenced metrics used in the task of open domain dialogue. The best performing model is highlighted in \textbf{bold}. $- w/o$ stands for the ablated model. \textbf{\textit{dy-agg}} denotes the aggregation on the dynamic graph, where \autoref{eq-H-enc} will exclude the input of $pseu_{G}$. \textbf{\textit{st-agg}} denotes the aggregation on static graph, where every element related to $\epsilon$ will be excluded, including the $q$ vector used in \textbf{\textit{dy-agg}}. \textbf{\textit{-w/o kg}} denotes the model learning the input without external graph knowledge (equivalent to vanilla BART). 
}
\end{table*}

\section{Experiment}
\subsection{Experimental Setup}
\paragraph{Dataset.} We process the dataset provided by~\citet{zhou2018commonsense} for the following experiments, where the train/val/test datasets are split into sizes of 3,384,185/20,000/10,000, respectively.\footnote{We follow prior work \cite{zhang-etal-2020-grounded} in using the original validation dataset as the test set for the convenience of comparison.} The statistics of the data are shown in \autoref{tab:data_stat}. From the table, it can be observed that the average statistics of entities, subgraphs and triples in these three splits are very close, implying that the data samples are fully shuffled to make the experiment fair.

\paragraph{Baselines.}
We select several competitive baselines for comparison, including: \textbf{Seq2seq} \cite{Seq2seq}, \textbf{MemNet} \cite{ghazvininejad2018knowledge}, \textbf{CopyNet} \cite{Copynet}, \textbf{UniLM} \cite{dong2019unified}, \textbf{BART} \cite{lewis-etal-2020-bart}, \textbf{CCM} \cite{zhou2018commonsense}, and \textbf{ConceptFlow} \cite{zhang-etal-2020-grounded}. 
In particular, UniLM and BART are SOTA pre-trained models for generation tasks, whilst ConceptFlow is the SOTA model for our task.\footnote{To our knowledge, ConceptFlow is the SOTA model for this task (where text and knowledge graphs are used as the input). There are some other similar works~\cite{yu-etal-2022-diversifying-content,wu2022improving} in commonsense dialogue generation, but they generate dialogues with additional documents or other kind of inputs. Due to the different input formats, they cannot be considered as baselines in our task.}

\paragraph{Evaluation Metrics.}
We adopt the metrics of BLEU \cite{Bleu}, NIST \cite{Nist}, METEOR \cite{Meteor}, Dist, and Ent \cite{zhang2018generating} for evaluation. BLEU, NIST, and METEOR are calculated between generated responses and golden responses, whilst Dist and Ent (calculating word distinction by $n$-grams) are calculated within generated responses. We also conduct further experiments to evaluate the efficiency and efficacy of incorporating external knowledge by counting the entities from the post used in the generated responses.

\subsection{Implementation Details}
Our framework is mainly implemented with Pytorch\footnote{\url{https://pytorch.org/}} and Pytorch-lightning, and we select BART~\cite{lewis-etal-2020-bart} as the base language model. We use a publicly available checkpoint\footnote{\url{https://huggingface.co/thu-coai/LongLM-base}} from Huggingface, and fine-tune it with our dynamic graph knowledge aggregation framework. The random seed is fixed to 42 for ease of reproducibility. Our language model has 12 attention heads and 6 hidden layers in each encoder and decoder, leading to a total of 157M parameters. The maximum sequence length is limited to 512; the \textit{batch size} is set to 64; and the \textit{learning rate} is 1e-4. We use  Adam~\cite{kingma2014adam} as the optimiser and set its parameter to 1e-8. The whole training process lasts for 5 \textit{epochs}. We train on an Nvidia RTX A100 GPU node, which has 120GB of system memory and 80GB of VRAM, and takes two days to train.

\begin{table}[t]
\centering \small
\resizebox{0.85\linewidth}{!}{
\begin{tabular}{l|cc|c}
\toprule
\textbf{Model} & \textbf{Dist-1} & \textbf{Dist-2} & \textbf{Ent-4}\\
\hline
\textbf{Seq2Seq}  & 0.0123 & 0.0525 & 7.665 \\
\textbf{MemNet}  & 0.0211 & 0.0931 & 8.418 \\
\textbf{CopyNet}  & 0.0223 & 0.0988 & 8.422 \\
\textbf{CCM}  & 0.0146 & 0.0643 & 7.847 \\
\textbf{UniLM}  & 0.0189 & 0.0755 & 9.599 \\
\textbf{Conceptflow} & 0.0223 & 0.1228 & 10.270 \\
\midrule
\textbf{SaBART (ours)} & 0.0598 & 0.2798 & 9.456 \\
\textbf{- w/o dy-agg} & 0.0607 & 0.2739 & 5.388  \\
\textbf{- w/o st-agg} & \textbf{0.0616} & \textbf{0.2816} & 9.916  \\
\textbf{- w/o kg} & 0.0055 & 0.1752 & \textbf{10.400}  \\
\bottomrule
\end{tabular}
}
\caption{\label{tab:unreference-metrics}
Automatic evaluation on unreferenced metrics.}
\end{table}

\begin{figure*}[t]
\centering
\includegraphics[width=0.98\linewidth]{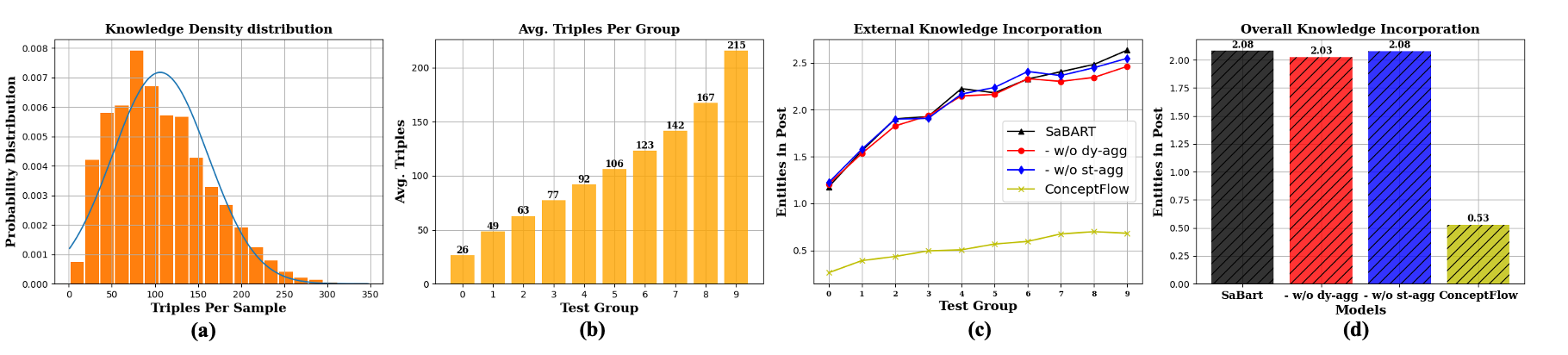}
\caption{The analysis of knowledge incorporation. The experiment groups datasets by the amount of external knowledge they contain. We perform evaluation with the knowledge amount increasing, so that the curve in (c) demonstrates the efficacy and efficiency of utilising external knowledge when generating responses. (d) is the average used knowledge amount in the whole dataset.} 
\label{fig:entity}
\end{figure*}

\begin{figure}[t]
\centering
\includegraphics[width=0.98\linewidth]{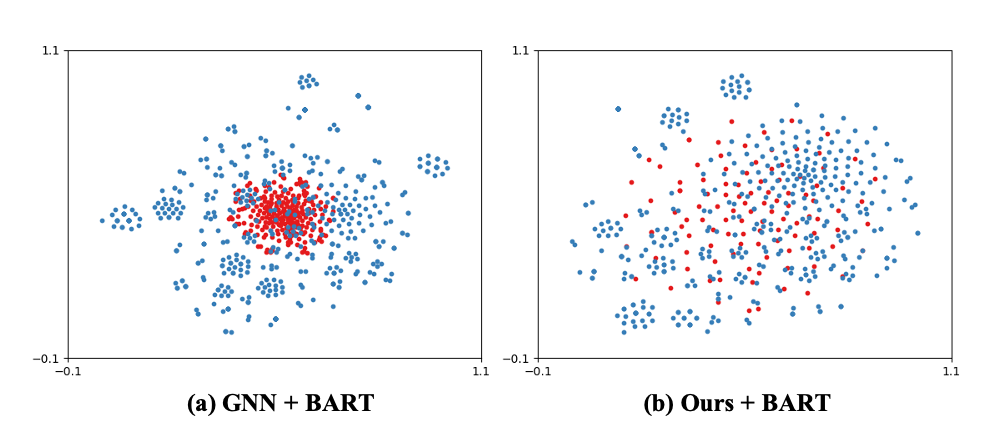}
\caption{The embeddings of entities (in red) and generic words (in blue) contained in the post projected by t-SNE. We extract these embeddings from 30 conversational pairs.} 
\label{fig:embeds}
\end{figure}

\subsection{Automatic Evaluation}
Table~\ref{tab:reference-metrics} shows reference-based automatic evaluation results, which demonstrate our proposed framework substantially outperforms all baselines on all referenced metrics. In comparison to the SOTA model ConceptFlow, our model doubles (e.g. BLEU-2, NIST-2, METEOR) or triples (e.g. BLEU-3, BLEU-4) performance on most metrics. Considering the performance of \textit{- w/o kg} (equivalent to vanilla BART), it can be inferred that the enhanced performance of our model is primarily attributable to 
incorporating external knowledge. 
Comparing to other GNN models (e.g. ConceptFlow and CCM), our model is superior in handling the heterogeneous features from the text and graph knowledge, leading to a better capturing of the global context contained in both the post and knowledge graph. In terms of unreferenced metrics, the results in \autoref{tab:unreference-metrics} also show that our model achieves a substantial improvement in diversity. It can be observed that our model performance on Dist-1 and Dist-2 are more than twice that of the SOTA model ConceptFlow. 
Our improvement on both unreferenced and referenced metrics further demonstrates that the gain comes from  incorporating knowledge to generate human-like responses, rather than metric-oriented training (i.e., no metric-oriented reward is used here).
In addition, the ablation results of \textit{- w/o dy-agg} and \textit{- w/o st-agg} also prove the hierarchical layers of graph knowledge aggregation benefit the semantic understanding of graph knowledge. The aggregation of static graph features forms the representation learning of lower-level semantics, whilst the dynamic aggregation contributes to the representation of higher-level semantics. Therefore, combining the two kinds of semantics leads to 
a substantial performance improvement on both referenced and unreferenced metrics.

\subsection{In-Depth Analysis}
Furthermore, we present two experiments to analyse whether the external knowledge is better exploited in our framework than the SOTA model (ConceptFlow), as well as why our framework learns representations more efficiently and effectively.

\paragraph{Performance of Knowledge Incorporation.}
The experimental results concerning the amount of knowledge used to generate responses are illustrated in \autoref{fig:entity}. Firstly, we group the test set by the number of post entities retrieved in the given post. The target is to analyse the robustness and efficiency of models as the knowledge content increases. \autoref{fig:entity}(a) indicates the probability distribution of the knowledge amount in each conversation pair, 3(b) shows the statistics of the grouped test set, 3(c) gives the curve illustrating how many retrieved knowledge items are finally used in generated responses, and 3(d) indicates that our model substantially outperforms the SOTA model by a large margin with respect to knowledge incorporation on the whole test set. It can be observed that with the proposed dynamic knowledge aggregation framework, the model tends to use more retrieved entities when generating a response. As the number of retrieved entities increases, the curve in (c) maintains a steady slope to incorporate entities, indicating that our model maintains the incorporation efficacy even with large amounts of knowledge as input. We argue that the robustness and efficiency of knowledge incorporation result from the globally aggregated features from the dynamically constructed graph with pseudo nodes, which avoids the information loss that the vanilla GNN models typically suffer. This also leads to our model outperforming other baseline models in generating high-quality responses.

\paragraph{Representations of Text and Graph Knowledge.} 
\autoref{fig:embeds} shows the representations of text and entities from the knowledge graph. We project the embeddings of the vanilla GNN models used in the baselines into two-dimensional points for visualisation. To compare the difference in embeddings from text and the knowledge graph, we normalise by mapping both embeddings to the range of $[0, 1]$. It can be observed that the entity embeddings of the baselines (shown in \autoref{fig:embeds}(a)) are concentrated in a circle no matter what post is given (i.e. blue points). This suggests that the GNN-learned embeddings present a biased representation for external knowledge, which may lead to difficulty in incorporating graph knowledge with the language model (which is trained on text corpora). In comparison, our framework unifies the representation hidden space of both text and graph knowledge, which makes the heterogeneous features have more shared space to fit the dataset. This mechanism makes the entity embeddings in our framework evenly spread among text words, thus it can be easily exploited by neural networks.

\subsection{Human Evaluation}
We present manual pair-wise comparisons to examine the \textit{appropriateness} (whether the response is appropriate in the context) and \textit{informativeness} (whether the response contains much information)  of the most competitive baseline (ConceptFlow), our model (SaBART), as well as two ablation models (\textit{- w/o dy-agg} and \textit{- w/o st-agg}). Three human evaluators are instructed to give their preferred response on 100 randomly sampled conversational pairs between each compared model. The results are reported in \autoref{tab:human evaluation}. 

When summarising the human annotation results, the final results are counted by majority voting. The ablated static aggregation and dynamic aggregation play different roles in feature incorporation, so the results of the corresponding ablation models are slightly lower than that of SaBART. On the other hand, the comparison with ConceptFlow demonstrates that our proposed model significantly outperforms the SOTA in terms of both appropriateness and informativeness, which is consistent with our observations in automatic evaluation.

\begin{table}[t]
\centering 
\resizebox{0.90\linewidth}{!}{
\begin{tabular}{r|lc|c}
\toprule[1pt]
\multirow{2}{*}{\textbf{Choice \%}} & \multicolumn{3}{c}{\textbf{SaBART \textit{vs} SaBART$_{\text{- w/o st-agg}}$}} \\
\cline{2-4} 
& \textbf{SaBART} & \textbf{\textit{- w/o st-agg}} & \textbf{$\mathit{Kappa}$} \\
\midrule
\textbf{\textit{App.}} & \textbf{62.3} & 37.7 & 0.363 \\
\textbf{\textit{Inf.}} & \textbf{60.0} & 40.0 & 0.418 \\

\midrule[1pt]
\multirow{2}{*}{\textbf{Choice \%}} & \multicolumn{3}{c}{\textbf{SaBART \textit{vs} SaBART$_{\text{- w/o dy-agg}}$ }} \\
\cline{2-4}
& \textbf{SaBART} & \textbf{\textit{- w/o dy-agg}} & \textbf{$\mathit{Kappa}$} \\
\midrule
\textbf{\textit{App.}} & \textbf{61.9} & 38.1 & 0.356\\
\textbf{\textit{Inf.}} & \textbf{65.0} & 35.0 & 0.438\\

\midrule[1pt]
\multirow{2}{*}{\textbf{Choice \%}} & \multicolumn{3}{c}{\textbf{SaBART \textit{vs} ConceptFlow}} \\
\cline{2-4}
& \textbf{SaBART} & \textbf{ConceptFlow} & \textbf{$\mathit{Kappa}$} \\
\midrule
\textbf{\textit{App.}} & \textbf{66.1} & 33.9 & 0.423 \\
\textbf{\textit{Inf.}} & \textbf{70.2} & 29.8 & 0.484 \\

\bottomrule
\end{tabular}
}
\caption{\label{tab:human evaluation}
Human Evaluation w.r.t.~\textit{appropriateness} and \textit{informativeness}. The score is the percentage that the proposed model wins against its competitor. $\mathit{Kappa}$ denotes Fleiss’ Kappa~\cite{fleiss1971measuring}, which indicates all of our evaluation annotations reach a fair or moderate agreement (meaning the two ablated models generate similar answers to \textbf{SaBART}). 
}
\end{table}

\subsection{Case Study}

\begin{table}[ht]
\centering
\resizebox{\linewidth}{!}{
\begin{tabular}{p{0.25\linewidth}|p{0.75\linewidth}}
\toprule
\textbf{Post \#1} & As an energy \textbf{drink} \textbf{consumer} who \textbf{hates} \textbf{tea}, this is \textbf{hilarious}. \\
\midrule
\textbf{SaBART} & I'm not a \textbf{tea} \textbf{drink} \textbf{consumer}, but i don't think it 's \textbf{hilarious}. \\
\midrule
\textbf{\textit{w/o st-agg}} & As a \textbf{consumer}, this is \textbf{hilarious}.  \\
\midrule
\textbf{\textit{w/o dy-agg}} & As a \textbf{consumer}, this is \textbf{hilarious}.   \\
\midrule
\textbf{ConceptFlow } &You 're a \textbf{fan} of \textbf{tea} !  \\
\midrule
\textbf{Post \#2} & \textbf{imo} the \textbf{game} deserves to be \textbf{played} on \textbf{pc}. Its not really suited for \textbf{controller}. I \textbf{played} it on \textbf{ps3} and sometimes it was just \textbf{frustrating}. \\
\midrule
\textbf{SaBART} & I've \textbf{played} it on \textbf{ps3}, and i have no \textbf{idea} what the \textbf{problem} is with it. \textbf{pc} is really \textbf{suitable} for this \textbf{game}. \\
\midrule
\textbf{\textit{w/o st-agg}} & It 's not suited for \textbf{pc}, but it doesn't have to be on a \textbf{ps3} .  \\
\midrule
\textbf{\textit{w/o dy-agg}} & I've \textbf{played} it on \textbf{ps3}, and it's not really suited for \textbf{pc}.   \\
\midrule
\textbf{ConceptFlow} & I \textbf{played} it for about a \textbf{year} and then i got to \textbf{play} it on my \textbf{ps3} and it was the only \textbf{game} i ever \textbf{played}. I didn't even know how to \textbf{play} it .  \\
\bottomrule
\end{tabular}
}
\caption{The case study for two conversational pairs extracted from the test dataset. Words in \textbf{bold} denote the graph entities in the posts (entity mentions) or in the responses (used knowledge).}
\label{tab:simple_case_study}
\end{table}

In \autoref{tab:simple_case_study}, we show the generated responses of the most competitive models (i.e., two ablated models and ConceptFlow). We select a short post and a long post that includes more graph knowledge to validate performance. It can be observed that in both cases, the models containing aggregation frameworks tend to compose their responses with more unique and relevant entities. When a short post is given, 
the knowledge can effectively help avoid generating overly simplistic utterances, resulting in a more appropriate, informative, and engaging response output. 
Given a long input, all models seem good at generating a long response. However, compared to SaBART, the responses generated by the baseline models are less expressive due to the sub-optimal incorporation 
of graph knowledge. For example, ConceptFlow uses four instances of ``play'' in the response, diluting the information it conveys, in addition to the content not being coherent or related to the post. In contrast, SaBART is able to better exploit the retrieved knowledge (e.g. ``frustrating'' in relation to ``problem''), which thus results in composing a more appropriate and informative response.

\section{Conclusion}
In this study, we propose a novel dynamic graph aggregation framework for the task of knowledge graph enhanced dialogue generation. We dynamically construct a graph with created pseudo nodes, and hierarchically aggregate graph knowledge to attain a global feature which better represents the topological structure and semantics contained in the knowledge graph. Due to the superior ability in leveraging the heterogeneous features of input text and graph knowledge, our framework can fill the semantic gap between the language model and knowledge graph and consequently generate an informative response. The extensive experiments demonstrate that our model significantly outperforms all baseline models, and can generate  more appropriate and informative responses utilising external graph knowledge.

\section*{Limitations}
This paper aims to investigate a more efficient and effective framework to incorporate the heterogeneous features of both text and graph knowledge. The extensive experiments demonstrate our framework has a superior performance in capturing semantics of input knowledge, thus beating all SOTA models. However, due to the time and resource limit, we could not conduct further experimentation to compare with promising frameworks in similar areas. In fact, we have observed some other techniques~\cite{tang-etal-2022-ngep, yu-etal-2022-diversifying-content, wu2022improving} may be beneficial to our study, but when considering the difficulty in applying them here (due to additional annotation and knowledge being required), we have to leave them to future work. We also cannot exclude some other factors which may affect performance. For example, we select BART as the base language model in this paper. In practical use, the latest language models (e.g. ChatGPT) may have better performance in this task. We have to leave the analysis of these factors to future study.

\section*{Ethics Statement}
We conduct the experiments based on an existing publicly available dataset from \citet{zhou2018commonsense} which is a large-scale dataset widely used to study commonsense dialogue generation, and we strictly follow the license and instructions. We also read and acknowledge the ACM Code of Ethnics and Professional Conduct.\footnote{\url{https://www.acm.org/code-of-ethics}} We take our professional responsibilities very seriously, and our study did not violate any ethical principles. Additionally, whilst our work concerns the incorporation of knowledge from knowledge graphs in dialogue systems, we acknowledge that the veracity and validity of the knowledge in such resources should be assessed in production, in order to avoid the perpetuation of misinformation.

\section*{Acknowledgements}
Chen Tang is supported by the China Scholarship Council (CSC) for his doctoral study (File No.202006120039). Tyler Loakman is supported by the Centre for Doctoral Training in Speech and Language Technologies (SLT) and their Applications funded by UK Research and Innovation [grant number EP/S023062/1].

\bibliography{acl2023}
\bibliographystyle{acl_natbib}

\appendix


\end{document}